\documentclass[lettersize,journal]{IEEEtran}
\usepackage{amsmath,amsfonts}
\usepackage{algorithmic}
\usepackage{algorithm}
\usepackage{array}
\usepackage[caption=false,font=normalsize,labelfont=sf,textfont=sf]{subfig}
\usepackage{textcomp}
\usepackage{stfloats}
\usepackage{url}
\usepackage{verbatim}
\usepackage{wrapfig}
\usepackage{graphicx}
\usepackage{cite}
\usepackage{xurl}
\usepackage[hidelinks]{hyperref}
\newcommand{\nop}[1]{}
\begin{document}

\title{Report for NSF Workshop on AI for Electronic Design Automation}

\author{
Deming Chen,\quad\thanks{Deming Chen is with University of Illinois Urbana-Champaign, Urbana, IL 61801 USA.}
\and
Vijay Ganesh,\quad\thanks{Vijay Ganesh and Yingyan Lin are with Georgia Institute of Technology, Atlanta, GA 30332 USA.}
\and
Weikai Li,\quad\thanks{Weikai Li, Jason Cong, and Yizhou Sun are with the University of California at Los Angeles, Los Angeles, CA 90095 USA.}
\and
Yingyan (Celine) Lin,\quad
\and
Yong Liu,\quad\thanks{Yong Liu is with Cadence Design Systems, Inc., San Jose, CA 95134 USA.}
\\
\and
Subhasish Mitra,\quad\thanks{Subhasish Mitra is with Stanford University, Stanford, CA 94305 USA.}
\and
David Z. Pan,\quad\thanks{David Z. Pan is with the University of Texas at Austin, Austin, TX 78712 USA.}
\and
Ruchir Puri,\quad\thanks{Ruchir Puri is with IBM, Poughkeepsie, NY 12601 USA.}
\and
Jason Cong*,\quad
\and
Yizhou Sun*
\thanks{* Corresponding authors: Jason Cong (cong@cs.ucla.edu) and Yizhou Sun (yzsun@cs.ucla.edu).}
}

\markboth{IEEE Circuits and Systems Magazine,~Volume:~26 Issue:~1, First Quarter~2026}{Shell \MakeLowercase{\textit{et al.}}: A Sample Article Using IEEEtran.cls for IEEE Journals}



\maketitle

\begingroup
\renewcommand{\thefootnote}{}  
\footnotetext{
\copyright~2026 IEEE. Personal use of this material is permitted.  Permission from IEEE must be obtained for all other uses, in any current or future media, including reprinting/republishing this material for advertising or promotional purposes, creating new collective works, for resale or redistribution to servers or lists, or reuse of any copyrighted component of this work in other works.

This is the accepted version. The published version is available at \url{https://ieeexplore.ieee.org/document/11466406}.
}
\endgroup

\begin{abstract}
This report distills the discussions and recommendations from the NSF Workshop on AI for Electronic Design Automation (EDA), held on December 10, 2024 in Vancouver alongside NeurIPS 2024. Bringing together experts across machine learning and EDA, the workshop examined how AI—spanning large language models (LLMs), graph neural networks (GNNs), reinforcement learning (RL), neurosymbolic methods, etc.—can facilitate EDA and shorten design turnaround. The workshop includes four themes: (1) AI for physical synthesis and design for manufacturing (DFM), discussing challenges in physical manufacturing process and potential AI applications; (2) AI for high-level and logic-level synthesis (HLS/LLS), covering pragma insertion, program transformation, RTL code generation, etc.; (3) AI toolbox for optimization and design, discussing frontier AI developments that could potentially be applied to EDA tasks; and (4) AI for test and verification, including LLM-assisted verification tools, ML-augmented SAT solving, security/reliability challenges, etc. The report recommends NSF to foster AI/EDA collaboration, invest in foundational AI for EDA, develop robust data infrastructures, promote scalable compute infrastructure, and invest in workforce development to democratize hardware design and enable next-generation hardware systems. The workshop information can be found on the website \url{https://ai4eda-workshop.github.io/}.
\end{abstract}


\section{Executive Summary}
\IEEEPARstart{T}{his} report summarizes the key discussions and recommendations from the NSF Workshop on AI for Electronic Design Automation (EDA), held alongside the NeurIPS 2024 conference in  Vancouver, Canada. The workshop brought together leading researchers and practitioners from both the machine learning (ML) and EDA communities to address the urgent need for AI-driven automation in chip design.

\textbf{The central challenge} in hardware design is the increasing complexity of modern chips. Domain-specific accelerators (DSAs) offer substantial performance and energy efficiency gains, but their design demands deep hardware expertise, leading to long turnaround times and hindering rapid innovation. AI techniques (such as Large Language Models), in combination with traditional EDA tools, have the potential to offer a transformative solution to democratize hardware design, making it domain-specific, cheaper, and far more effective.

The workshop explored four key themes:

\begin{itemize}
\item \textbf{AI for physical synthesis and Design for Manufacturing (DFM):} This theme addressed the challenges in translating a chip design into a manufacturable product, including multi-objective optimization, complex constraints, and long turnaround times. Current AI efforts leverage Reinforcement Learning (RL) for placement, surrogate models for analysis, LLMs for rule creation, and generative AI for design acceleration. Key showstoppers include generalization issues across designs and nodes, scalability limitations, data quality and availability, and the significant computational resources required for AI models. Opportunities lie in fostering cross-community collaboration, developing open-source frameworks and datasets, and exploring hybrid AI approaches that combine deep learning with traditional methods. Recommendations to NSF include supporting foundational AI research for EDA, encouraging open-source initiatives, promoting scalable compute infrastructure, and investing in workforce development.

\IEEEpubidadjcol    
\item \textbf{AI for High-Level and Logic-Level Synthesis (HLS and LLS):} This section focused on automating the translation of high-level behavioral descriptions into gate-level specifications. The core problem is the manual, error-prone, and time-consuming nature of traditional synthesis, exacerbated by the labor disparity in hardware design. AI can automate design decisions, predict performance outcomes, and enhance design space exploration. Existing efforts include HLS coupled with deep learning optimization (HLS+DL), using Graph Neural Networks (GNNs) and LLMs for performance prediction and pragma insertion, and more ambitious natural language to RTL (NL2RTL) generation. Showstoppers include the difficulty in adapting AI models to toolchain changes and new designs (domain shift), and the critical need for high-quality, labeled training data. Significant opportunities exist for the AI community to tackle novel problems in multi-modality fusion, domain/task transfer, and optimization, while the EDA community can leverage AI for accelerated pipelines, synthetic data generation, and black-box optimization.

\item \textbf{AI toolbox for optimization and design:} This section focused on challenges raised to AI from chip design. Existing AI techniques rely heavily on well-maintained offline data, a clearly defined objective function, and an offline evaluation. Key recommendations for NSF include the following. Generalization capability and uncertainty estimation become even more critical in chip design. Reinforcement learning and black-box optimization can lead to more effective and efficient search, which is beyond the traditional prediction tasks in general ML. LLM agents could serve as a very powerful tool by automatically putting long pipelines together and getting end-to-end solutions from data collection and tool calling to verification.

\item \textbf{AI for test and verification:} This session explored the transformative role of artificial intelligence in test and verification within the domain of EDA. Key discussions included methods for learning robust design representations that generalize across multiple hardware design tasks, enhancing the adaptability and efficiency of AI-driven solutions. A focal point was an LLM-aided hardware design framework, which is uniquely enabled by G-QED (Generalized Quick Error Detection), a new verification technique for rapidly identifying logic bugs during chip verification. The workshop also addressed the evolving landscape of software development in the era of LLMs, emphasizing their application in fuzz testing and the novel use of LLMs for generating effective fuzzing programs. Another topic covered was security vulnerabilities of AI-based techniques, as they are universally being employed in various testing tasks, and their unreliability can lead to nefarious system breaches. Finally, presenters showcased AI-based optimization strategies to efficiently solve complex SAT (satisfiability) verification problems, highlighting a path toward more scalable and accurate hardware verification processes. Collectively, the session emphasized the synergy between AI and EDA to enable faster, more intelligent design and design verification flows across the design lifecycle.
\end{itemize}

\textbf{Overall, the workshop highlighted the immense potential of AI to revolutionize chip design.} Realizing this potential requires a concerted effort to overcome challenges related to data availability, generalization, scalability, and integration. One common theme that came up was how to find optimized combinations of AI and non-AI approaches for large benefits -- a truly cross-disciplinary approach to exploit the positives of each and avoid their negatives. Key recommendations for the NSF and the broader research community include:

\begin{itemize}
\item \textbf{Fostering cross-disciplinary collaboration} through shared datasets, benchmarks, open-source tools, and joint research programs.
\item \textbf{Investing in foundational AI research} tailored for EDA, with a focus on explainability, scalability, and robustness.
\item \textbf{Developing robust data infrastructures} to efficiently extract and manage EDA data, alongside establishing diverse and large-scale benchmark datasets.
\item \textbf{Promoting scalable compute infrastructure} and encouraging the development of hybrid AI approaches that combine the strengths of AI with traditional EDA methodologies.
\item \textbf{Investing in workforce development} to train a new generation of professionals at the intersection of AI and EDA.
\end{itemize}

By addressing these critical areas, the workshop aims to accelerate the adoption of AI in chip design, ultimately democratizing hardware innovation and enabling the creation of next-generation energy-efficient computing systems.

\section{Introduction and Background}

\noindent In the past few years, domain-specific accelerators (DSAs), such as Google's Tensor Processing Unit (TPU), have been shown to offer significant performance and energy efficiency over general-purpose CPUs. However, DSAs require deep hardware knowledge to achieve high performance and robustness. Unfortunately, there are far fewer hardware designers than software developers. Leveraging AI techniques to further automate chip design becomes the key to meeting the need for rapid change in software development, which is also critical for democratizing hardware design and creating next-generation energy-efficient hardware. This workshop aims to bring researchers and practitioners from both machine learning and EDA communities to discuss how AI can address challenges in different stages of hardware design, promoting open benchmark datasets and open discussions to revolutionize chip design.

The workshop had 20 invited speakers and panelists covering four themes, and attracted 130+ attendees, on a wide variety of methods and problems in this area. The workshop covers four themes, including three key areas of EDA and the AI toolbox:

\begin{itemize}
\item AI for physical synthesis and design for manufacturing
\item AI for high-level and logic-level synthesis
\item AI toolbox for optimization and design
\item AI for test and verification
\end{itemize}

The workshop includes a wide range of topics, including but not limited to:
\begin{itemize}
\item Graph Neural Networks for Performance, Power, and Area (PPA) prediction and optimization for all stages of the EDA pipeline, including HLS, RTL/logic synthesis, and physical designs;
\item LLM-based code analysis for PPA prediction and optimization at all stages of the EDA pipeline;
\item Multi-modality ML models (e.g., LLM + GNN) for PPA prediction and optimization at all stages of the EDA pipeline;
\item Domain and task transfer for HLS performance prediction with new kernels and new versions of EDA tools at different stages;
\item ML methods to optimize circuit aging and reliability;
\item ML for Design Technology Co-Optimization (DTCO);
\item ML for analog, mixed-signal, and RF IC designs;
\item Reinforcement learning for Design Space Exploration (DSE);
\item LLM-based design generation;
\item Interplay between AI-enabled design and verification;
\item Active learning and importance sampling of design points;
\item AI for compiler and code transformation;
\item Benchmark datasets.
\end{itemize}

Based on the three key areas of EDA (AI for physical synthesis and design for manufacturing, AI for high-level and logic-level synthesis, and AI for test and verification), we formed round-table discussions to identify key challenges, SOTA approaches, and future directions. Note that Theme 3 (AI toolbox for optimization and design) speakers and panelists participated in the other three themes’ discussions.

\section{AI for Physical Synthesis and Design for Manufacturing}

\subsection{Problem Description and Challenges}

\noindent Physical synthesis and Design for Manufacturing (DFM) are essential stages in Electronic Design Automation (EDA), ensuring that a chip design can be efficiently and reliably produced. Modern chip design, with billions of transistors and advanced process nodes like 3nm and 2nm, presents significant challenges to these processes. Traditional manual and iterative approaches struggle to cope with the immense complexity and time constraints of modern chip development. Key challenges in this area are:

\begin{itemize}
\item Multiple objective optimizations: Physical design, DFM, and analog designs are multiple objective optimization problems. 
\item Complex constraints: Physical synthesis involves numerous constraints such as timing, power, density, and manufacturability, which grow exponentially with scaling to advanced nodes like 5nm, 3nm, and 2nm.
\item DRC complexity: Design Rule Checking (DRC) involves thousands of rules, which must be checked and optimized. Many rules are negotiable with customers, further complicating the process.
\item Combinatorial optimization: Placement, routing, and gate sizing are inherently combinatorial problems, often requiring significant computational resources to achieve near-optimal solutions.
\item Analog and RFIC designs: While digital designs have well-established EDA abstractions and tools, analog and RF IC designs are still heavily manual and not scalable.
\item Long turnaround times: Traditional methods result in weeks of delay for large designs due to iterative processes in routing, simulation, and verification.
\item Manufacturing variations: Ensuring yield and robustness against process variations is increasingly difficult at advanced nodes, where small discrepancies can lead to significant defects.
\end{itemize}

AI/EDA has the potential to overcome these challenges.

\subsection{Summary of Existing Efforts and Showstoppers}

\noindent \textbf{Existing efforts}:

\begin{itemize}
\item Reinforcement Learning (RL): Used for macro placement tasks~\cite{graph_placement}, RL has shown some success in improving placement quality. RL has also been shown for analog transistor sizing to meet tough design specifications~\cite{DNN-Opt}.
\item Surrogate models for design analysis: Faster and approximate models replace traditional analytical analysis for key design metrics such as timing, power, routability, etc., accelerating optimizations~\cite{pre_routing_slack_prediction}.
\item LLM for rule creation: Large language models (LLMs) are being applied to automate DRC code writing and verification~\cite{DRC-Coder}.
\item Generative AI for physical design: Leverage generative AI models trained on optimized data to achieve a significant boost over traditional optimization methods, e.g.,~\cite{DREAM-GAN}.
\item Gradient-based method for physical design: model physical optimization problems as differentiable problems and optimize with gradient-based methods using deep learning frameworks on GPU~\cite{DREAMPlace,INSTA}.
\item Machine Learning (ML) for placement and routing, e.g., predictive tools assist in congestion analysis and timing closure~\cite{congestion_aware_placer}.
\item AI in DFM analysis: Techniques for hotspot detection~\cite{lithographic_hotspot_detection}, printed image~\cite{LithoGAN} and process window predictions, and yield optimization to enhance design robustness.
\item Simulation acceleration: AI-driven surrogate models provide near-real-time predictions~\cite{Neurolight}, speeding up simulation processes.
\item Generative AI for analog and RF design: Emerging tools like AnalogCoder~\cite{lai2025analogcoder} and PulseRF~\cite{CHAE_ICCAD24} enable faster design cycles and optimize performance.
\item AI in early feedback loops: AI-powered tools can provide faster feedback during layout stages, real-time DRC (Design Rule Check) violation detection~\cite{Design_rule_checking}, supporting informed decision-making and improving productivity.
\end{itemize}

\noindent \textbf{Showstoppers}:
\begin{itemize}
\item Generalization: AI models often fail to adapt to new designs or nodes or designs that have different characteristics from the training data.
\item Scalability: Limited scalability of current solutions for detailed routing.
\item Data quality: Garbage-in, garbage-out issues in AI training data often limit the effectiveness of predictive models.
\item Data availability: EDA and design data are limited and often not in the open domain. It is also difficult to extract data from existing design databases efficiently. 
\item Computational resource demands: AI-based methods can require significant computing power, making them inaccessible to smaller organizations.
\item Integration challenges: Lack of seamless integration across EDA tools hinders the realization of end-to-end AI-driven workflows.
\end{itemize}

\subsection{Opportunities for AI/EDA Collaboration}

\begin{itemize}
\item Facilitating cross-community collaboration:
\begin{itemize}
\item Open-source initiatives for developing shared datasets, tools, and benchmarks.
\item Community-wide competitions to foster collaboration and innovation in both AI and EDA communities.
\item Standardizing AI integration for interoperability across tools and workflows.
\item Joint research programs to address real-world challenges effectively.
\item Joint summer schools / bootcamps on AI + EDA to cultivate the next generation of workforce.
\end{itemize}

\item Advanced modeling techniques: Collaborative research to improve multi-fidelity models for early-stage optimization,  along with innovations in multi-modality AI.
\item Open-source frameworks: Develop public benchmarks and datasets for AI-driven design, placement, and routing to enable reproducibility and innovation. Develop open-source frameworks to foster collaboration between AI and EDA communities, enabling the development, evolution, and contribution of integrated end-to-end solutions for AI+EDA in a collaborative manner.
\item Hybrid AI approaches: Combine deep learning with traditional algorithmic approaches or heuristic-based methods to tackle hard optimization problems in EDA.
\item Human-AI collaboration: A human-in-the-loop model enables seamless collaboration between AI tools and designers, allowing AI to handle repetitive tasks and complex simulations while leveraging human expertise and domain knowledge to guide, refine, and advance the development of AI solutions.
\end{itemize}

\subsection{Future Directions and Recommendations to NSF}
\noindent \textbf{Future directions} include, but are not limited to:

\begin{itemize}
\item Improve the generalization and scalability of AI models to effectively manage the complexities of modern chip design. Exploring the use of synthetic data and federated learning to address data scarcity and intellectual property concerns.
\item Generative and gradient-based AI: Combining generative AI with gradient-based methods shows promise in optimizing physical design while balancing performance, power, and area (PPA). 
\item Agentic systems to automate physical design tasks, including design flow generation and physical design debug, etc.
\item Multi-modal models to understand diverse data modality in the physical design, such as images, netlist/graph, RTL/text. 
\item Faster design cycles: Focus on streamlining manual tasks, enabling faster iteration cycles, improving overall throughput.
\item End-to-end automation: Focus on creating systems capable of automating chip design from architecture specification to GDSII.
\item Cross-layer co-optimization: Optimize across multiple design stages, including logic synthesis, placement, and routing.
\item Human-in-the-loop systems to guide AI-driven design processes. Enhanced explainability for building trust and debugging AI decisions.
\item Multi-fidelity models to bridge early-stage design and late-stage validation.
\item AI-enhanced DFM: Incorporate ML-based surrogate models to rapidly predict and optimize manufacturability and yield during the design phase.
\item Parallel processing: Leverage GPU/TPU-accelerated frameworks to significantly reduce turnaround times for large-scale optimization problems.
\item Multimodal generative AI: Expand the use of multi-modal transformers to leverage all the different modalities of data available in the DFM. 
\item 3D IC integration: Explore AI-driven design and optimization techniques for 3D ICs, addressing challenges in vertical interconnect planning, thermal management, and multi-die co-design.
\item Data infrastructure for EDA: open-source and commercial data infrastructures to extract EDA data efficiently from open-source and commercial EDA platforms. 
\item Benchmarks for AI on EDA: diverse sets of benchmarks for various sets of EDA problems, such as placement, optimization, routing, CTS, as well as analog/RF optimization and layout problems. 
\end{itemize}

\noindent Based on the discussions, we propose the following \textbf{recommendations to NSF}:

\begin{itemize}
\item Support foundational AI research for EDA: Focus on AI techniques tailored for EDA with explainability, scalability, generalization, and robustness.
\item Encourage cross-disciplinary collaboration: Facilitate partnerships through workshops, competitions, and funding programs.
\item Encourage open-source collaboration: Create incentives for industry and academia to develop and share datasets, benchmarks, and AI models.
\item Promote scalable compute infrastructure: Provide grants for access to high-performance computing environments needed for training large AI models.
\item Drive innovation in cross-layer Design Technology Co-Optimization (DTCO): Establish specific programs to fund digital-twin research and AI-driven DTCO tools.
\item Invest in workforce development: Enhance education programs to train professionals in the intersection of AI and EDA to foster a skilled workforce.
\item Support benchmark and dataset efforts: solicit and support benchmarks and datasets for AI for EDA tasks; propose grand challenges and contests on AI for EDA.
\end{itemize}

\section{AI for High-Level and Logic-Level Synthesis}
\label{sec:HLS_LLS}

\subsection{Problem Description and Challenges}

\noindent \textbf{Problem description:} High-Level Synthesis (HLS) and Logic-Level Synthesis are pivotal stages in hardware design, translating high-level behavioral descriptions into gate-level specifications. However, these processes are traditionally manual, error-prone, and immensely time-consuming, requiring deep domain expertise that is in short supply. AI, particularly techniques like Machine Learning (ML) and Large Language Models (LLMs), offers the potential to automate and optimize these synthesis stages, which could significantly speed up the design process while also enhancing the performance, power, and area (PPA) efficiency of the resulting circuits. This session and topic focus aims to delve into the integration of AI in HLS and logic-level synthesis, exploring how AI can address the inherent challenges in these stages. Specific focus areas will include:

\begin{itemize}
\item Automating design decisions: AI, particularly through the use of machine learning models like GNNs and LLMs, can automate the insertion of pragmas and optimize other design decisions that traditionally require deep domain expertise.
\item Predicting performance outcomes: AI models can predict chip performance early in the design phase, allowing for rapid iterative improvements without the need for full synthesis runs.
\item Enhancing design space exploration: AI can significantly expand and efficiently explore the design space to identify optimal design configurations, far beyond what is feasible with traditional methods.
\end{itemize}

\textbf{Challenges:} The integration of AI into high-level and logic-level synthesis (HLS and LLS) is facing several critical challenges that stem from both workforce disparities and technical complexities:

\begin{itemize}
\item Labor disparity in hardware design: The significant gap between the number of software developers (approximately 1,795,300) and hardware designers (78,100) highlights a labor crisis in hardware design. This disparity is problematic as it limits the number of skilled personnel available to meet the increasing demand for specialized hardware. To counter this, research has been directed towards empowering software developers to participate in chip design, aiming to democratize and accelerate the hardware design process.
\item Data challenges: AI-driven approaches in HLS and LLS heavily rely on high-quality, labeled data, which is scarce, proprietary, sensitive, and costly to produce. The challenge lies in developing data-sharing mechanisms that protect the proprietary aspects of designs while providing sufficient data to train effective AI models. This balance is crucial for advancing AI capabilities in chip design without compromising intellectual property. 
\item Integration with existing EDA tools: Integrating AI tools into the existing complex and often outdated EDA software stacks presents another hurdle. AI models must seamlessly interact with these tools, requiring extensive engineering efforts and a profound understanding of both software and hardware domains. This integration is essential for practical application, leveraging AI innovations effectively in the synthesis process.
\item Trust and reliability: There is a fundamental need for trust in AI-generated designs, which must adhere to stringent industry standards. Building this trust involves developing comprehensive validation and verification processes that are both thorough and efficient. However, these processes can be resource-intensive and challenging to establish, posing a significant barrier to the routine use of AI in critical design tasks.
\end{itemize}

These challenges underscore the complexities of applying AI to revolutionize HLS and LLS. Addressing these issues requires not only technological innovations but also strategic collaborations across the industry to enhance data availability, tool integration, and trust in AI-generated outputs.

\subsection{Summary of Existing Efforts and Showstoppers}

\noindent \textbf{High-level synthesis.} Currently, there are two types of AI-assisted efforts in design creation. One starts with a high-level specification, such as C/C++ or even Python, and gets refined all the way down to RTL code with the help of high-level synthesis (HLS) coupled with deep learning optimization (HLS+DL). The other effort is more ambitious, starting with natural languages and generating RTL code directly (NL2RTL).

\textbf{HLS+DL.} For the class of HLS+DL approaches in the first category, an early work is GNN-DSE~\cite{GNN-DSE}, which represents the input C/C++ programs as a control and dataflow graph (CDFG), and uses graph neural networks to build a predictive model so that circuit performance can be predicted in milliseconds without running the HLS tool. GNN-DSE is trained using a database of HLS designs generated by an expert-like system called AutoDSE~\cite{AutoDSE}. This model is enhanced to HARP~\cite{HARP}, which adds hierarchical modeling to deal with the long-dependency issues in the input program. HARP also models pragmas as function transformations. With the rapid advance in LLM, a subsequent work, ProgSG~\cite{ProgSG}, combines GNN and LLM with cross-modality attention.

One serious challenge of any of these predictive models is to track the changes of the toolchain (e.g., the HLS tool) they plan to model. This problem is addressed by Active-CEM~\cite{Active-CEM}, which carries out a task transfer learning scheme that leverages a model-based explorer designed to adapt efficiently to changes in toolchains. This approach optimizes sample efficiency by identifying high-quality design configurations under a new toolchain without requiring extensive re-evaluation. It further refines the methodology by incorporating toolchain-invariant modeling.

Another major challenge arises when the trained model is applied to new designs, as the substantial domain shift often results in unsatisfactory performance. One attempt to address this problem is to use a hierarchical MoE~\cite{Hierarchical_MoE}, which proposes a more domain-generalizable model structure. Different expert networks can learn to deal with different regions in the representation space, and they can utilize similar patterns between the old kernels and new kernels. The MoE can be applied at the basic block level, loop/function level, and the graph level, with the assumption that even for a new kernel, it will share common basic blocks and loop structures with some training examples, although the composition of these basic structures can be different. This approach showed some promising results for adapting to new domains. Nevertheless, more effective and robust methods are needed to transfer learning to new designs.

So far, these HLS+DL methods focus on pragma insertion, which is very important in defining the microarchitecture of the chip, and rely on the underlying Merlin compiler~\cite{Merlin} to carry out limited program transformation, such as loop strip mining and memory access coalescing. However, more substantial transformation of the original program, e.g., with loop ordering and distribution, may achieve considerably better results, as demonstrated in several non-ML-based approaches, including ScaleHLS~\cite{ScaleHLS}, HIDA~\cite{Hida}, Allo~\cite{Allo}, StreamHLS~\cite{Stream-HLS}, and Sisyphus~\cite{Sisyphus}. However, most of these approaches are limited to affine programs, so that some analytical models can be derived for performance prediction. It will be interesting to incorporate these methods and expand the design space of the existing HLS+DL methods to include program transformation for non-affine programs.

\textbf{NL2RTL.} For the NL2RTL efforts in the second category, which begin with natural language inputs to generate RTL code, early explorations like GPT4AIGChip~\cite{GPT4AIGChip} demonstrate the feasibility of this approach. GPT4AIGChip processes natural language specifications that define target design specifications and resource constraints, such as latency and DSPs, to produce RTL designs according to the provided demonstration examples of AI accelerators. Initially validated in the context of CNN accelerators, this method illustrates how advanced language models can potentially simplify and automate complex design tasks.

Motivated by the effectiveness of merely providing a few demonstration examples in GPT4AIGChip, MG-Verilog~\cite{MG-Verilog} introduces an approach to enhancing LLM-assisted Verilog generation by providing a dataset of 11k that includes multi-grained, detailed descriptions of hardware functions alongside their corresponding Verilog code. This dataset aims to improve the training and operational efficacy of language models in hardware design by offering diverse levels of description detail, from high-level summaries to intricate, line-by-line annotations.

LLM4SVA [under review] presented a further development in this direction, targeting the automated generation of SystemVerilog Assertions (SVAs) from high-level natural language descriptions. This framework aims to bridge the verification gap that arises with NL2RTL approaches by providing an LLM-based method to ensure that the RTL code/design adheres to its specifications, thereby enhancing reliability and reducing manual verification efforts.

The NL2RTL direction still faces significant challenges, particularly in terms of accuracy and the faithful translation of functional requirements into hardware designs. To address these concerns, continuous advancements in language model training methodologies and hardware architecture-specific adaptations are being explored. The integration of domain-specific knowledge into the training process of LLMs is crucial for achieving higher precision and functional correctness in generated RTL code.

Moreover, the verification of designs generated through NL2RTL approaches remains a critical hurdle. It requires innovative solutions to ensure that the designs not only meet performance metrics but also adhere strictly to safety and regulatory standards. The development of integrated tools that can automatically generate and verify RTL from natural language will be key to the maturation and acceptance of NL2RTL technologies in mainstream hardware design workflows.

Overall, while still in the experimental phase, the NL2RTL effort holds the promise of significantly streamlining the design process, making it faster and more accessible to designers without deep expertise in traditional hardware description languages. These developments, paired with advancements in AI and machine learning, are poised to redefine the landscape of digital design in the coming years.

Having high-quality training data is important for the success of the approaches in both categories. The HLsyn benchmarks~\cite{HLSyn} provided 80,000 HLS programs with performance and resource utilization data, which was very helpful for several subsequent research (e.g.,~\cite{ProgSG,Active-CEM,Hierarchical_MoE}). The Chrysalis dataset~\cite{Chrysalis} (symbolizing a phase in the life cycle where a ‘bug’ transforms into a ‘butterfly’) includes an extensive collection of over 1,500 HLS designs, annotated with 19 distinct bug types through LLM-based bug injection techniques. A contest was organized at ICCAD 2024 to collect benchmarks for NL2RTL efforts~\cite{LLM4HWDesign}. More efforts are needed in this direction so that we can reach the training data size of ImageNet scale.

\textbf{Logic-level synthesis.} On the logic-level synthesis front, the following ML-driven works have been studied, showcasing promising strides in addressing challenges such as design optimization, resource usage estimation, and synthesis automation. Leveraging ML techniques like deep neural networks (DNNs), reinforcement learning (RL), and symbolic reasoning, these works have advanced the quality of results (QoR) and efficiency in logic synthesis of digital chip design.

\begin{itemize}
\item Accurate estimation of design metrics with ML: High-Level Synthesis (HLS) often encounters discrepancies between estimated and post-implementation metrics. To address this, Pyramid~\cite{Pyramid} introduces an ensemble ML model trained on diverse benchmarks that predicts timing and resource usage with over 95\% accuracy. Thus, it reduces the need for exhaustive implementation and enables early assessment of design trade-offs. 

\item Dataset generation for ML tasks in logic synthesis: Datasets are foundational for ML models in logic synthesis. OpenLS-DGF~\cite{OpenLS-DGF} generates over 966,000 Boolean circuits in Verilog and GraphML formats, supporting various ML tasks like circuit classification and QoR prediction, advancing ML applications in logic synthesis.

\item Automation through reinforcement learning: The exploration of vast design optimization spaces benefits significantly from RL-based approaches:
\begin{itemize}
\item DRiLLS~\cite{DRiLLS} applies the Advantage Actor-Critic (A2C) algorithm to navigate optimization spaces autonomously. It achieves a 13\% average QoR improvement on EPFL benchmarks by minimizing area under timing constraints.
\item AISYN~\cite{AISYN} extends this concept by overcoming local minima in classical synthesis methods. Leveraging RL, AISYN optimizes delay, area, and power, achieving up to a 69.3\% reduction in cell area compared to traditional methods. These works highlight RL's potential to outperform human-driven optimization processes.
\end{itemize}

\item Hybrid approaches to logic synthesis: LSOracle~\cite{LSOracle} integrates traditional synthesis techniques with AI by combining And-Inverter Graph (AIG) and Majority-Inverter Graph (MIG) optimizers. Utilizing DNNs to classify circuit partitions, it applies the most suitable optimizer, achieving a 6.87\% improvement in area-delay product for 7nm ASIC designs. This hybrid framework bridges the gap between automation and human expertise.

\item Logic minimization with synergistic learning: The INVICTUS~\cite{Invictus} framework addresses logic minimization by automating optimization sequence generation through model-based offline RL. By combining RL and search methods, INVICTUS synthesizes Boolean circuits, achieving up to a 30\% improvement in area-delay product (ADP) and a 6.3× reduction in runtime. This scalability positions it as a powerful solution for various circuit benchmarks.

\item Symbolic reasoning for functional understanding: Gamora~\cite{Gamora} applies symbolic reasoning at scale, using graph neural networks (GNNs) and GPU acceleration to extract high-level functional blocks from gate-level netlists. Supporting large-scale Boolean networks, Gamora enhances the analysis and understanding of complex designs, opening avenues for advanced functional insights.
\end{itemize}

These ML-driven advancements highlight the transformative potential of AI in logic synthesis, enabling more efficient, accurate, and automated workflows. From precise metric estimation and adaptive dataset generation to RL-based optimization and hybrid frameworks, these works represent significant progress. However, challenges remain in generalizing across diverse designs, bridging theory with industrial adoption, and scaling for emerging technologies. Part of the challenge stems from the lack of ML-friendly representations of EDA data, as well as the inherent complexity of integrating logical, geometric, and numeric reasoning into ML.

\subsection{Opportunities for AI/EDA Collaboration}

\noindent \textbf{Opportunities for AI community:}
\begin{itemize}
\item Logic synthesis and high-level synthesis (HLS) provide unique new problems to the AI community. AI researchers prefer problems with clear inputs and outputs, which can be fairly evaluated and benchmarked. These problems have common challenges seen by other AI applications, such as multi-modality fusion, domain transfer, task transfer, and label scarcity. These problems have also raised unique challenges, such as delayed reward, expensive to run evaluation, and an extremely large search space.
\item AI for optimization can find a new arena in this area.
\item AI for simulation would be an essential technique to help accelerate the process of collecting (early) feedback for designs.
\item How to effectively leverage pre-trained foundation models to turn natural language into specs and then Verilog will be interesting.
\end{itemize}

\noindent \textbf{Opportunities for EDA community:}
\begin{itemize}
\item AI tools can significantly accelerate the pipelines and simplify the workflow involved in EDA.
\item AI for simulation can be used to generate large-scale synthetic datasets.
\item GenAIs and black-box optimization can be used to propose high-quality design points without the necessity to brute-force search.
\item LLM-based agents can work with human experts.
\end{itemize}

\noindent \textbf{How to facilitate cross-community collaboration:}
\begin{itemize}
\item Build more benchmark datasets and set up leaderboards.
\item Advertise more in the AI community and host contests. A good venue could be the NeurIPS contest track. The AI community uses social media significantly. 
\item Continue AI4EDA workshops in ML conferences.
\item Introduce AI topics into EDA courses and introduce EDA problems into AI courses.
\end{itemize}

\subsection{Future Directions and Recommendations to NSF}

\noindent \textbf{Future Directions on EDA:}
\begin{itemize}
\item Agentic-EDA: Agents are enabling an exciting new paradigm in the evolution of AI systems - systems that build intelligence at inference/real-time vs training time only, and are shifting the system behavior from feedforward systems (prompt-LLM-output) to feedback systems (Think/Plan – Act with Tools – Observe/Reflect – Loop – Re-Think/Plan – Act with Tools – Observe/Reflect – …). This offers the opportunity of a significant productivity boost for the chip designers as the level of automation can move from a single sub-task to a complex end-to-end task, like fixing a bug/timing/area/noise/power issue in a design with agentic workflows. A multi-agent system that combines various sub-agents specializing in different aspects of EDA with human-in-the-loop is the future of agentic-EDA.

\item Carefully engineered AI/ML tools: To help chip designs, the following directions are promising: combining multiple ML models, such as GNN and LLM, for more comprehensive representations of the circuit; capturing design hierarchy and pragma transformations and multi-modality attention; task transfer learning with active learning and domain transfer learning with MoEs, etc. In addition, ML may be useful when (1) an IC is implemented many times after small changes (then we have a large dataset for training); (2) faster prototyping of combinatorial solvers; (3) optimization achieved through Reinforcement Learning, even when the optimization objective is not available in a closed form.

\item Extracting and learning of information: Streamline the process of extracting/learning of information from existing designs and optimization runs, and apply such knowledge/expertise to speed up and enhance future designs. This strategy can be applied to virtually any aspect of the EDA design and verification space, including Logic and High-Level Synthesis. Examples include: (1) Design to design: learn from existing designs to predict PPA for new designs; (2) Run to run: serve as congestion/timing/area predictor during optimization; (3) Learning and predicting the desirable Technology / Library configurations.

\item Predicting key performance metrics: New ML models trained to predict key performance metrics from high-level code or intermediate representations for circuits, reducing dependency on slow simulation/implementation, which can help DSE and identify high-quality design points. This can be high-level architecture decisions at HLS and suitable optimization at the logic level.
\end{itemize}

\noindent \textbf{Other Promising Directions:}
\begin{itemize}
\item Prioritize optimization moves (e.g., retiming, logic restructuring, sizing, placement), based on learned performance trade-offs.
\item Generate new datapath elements optimized for specific metrics, instead of relying on library templates.
\item Enhance “Lint” verification and predict problematic code, possible congestion, and timing problems based on learned experience from previous designs.
\item Assist in interpreting implementation error messages and suggesting fixes.
\item Focus on predictability, stability, repeatability, and generalizability (over-fitting results to a particular data set might limit QoR).
\end{itemize}

\noindent \textbf{Break the Data Barrier in EDA}
\begin{itemize}
\item A framework to develop a truly open EDA LLM with a taxonomy, synthetic data generation, and iterative large-scale alignment/tuning approach (e.g., IBM’s open community and open source InstructLab approach) that enables LLM contributions of knowledge and skills from the community. This will help us overcome the data bottleneck in EDA, which has hampered the progress of AI in the Chip Design Automation domain.
\item Large-scale RTL code generation: Use high-performance HLS flows to generate a large amount of high-quality RTL code, e.g., leveraging the HIDA flow~\cite{Hida} and Allo flow~\cite{Allo}. 
\item Utilizing low-quality data for training: Semiconductor design teams create huge amounts of “low quality” data on a daily basis: implementation tools log files and reports; tool settings and run options; PPA info at various implementation stages, etc. It will be helpful to create a process to bring this data to a level that is useful for training purposes.
\end{itemize}

\noindent \textbf{Recommendations to NSF:}
\begin{itemize}
\item Promote benchmarking data and tasks.
\item Promote open frameworks and platforms.
\item Encourage new AI tools, such as LLM-based agents.
\item Encourage novel multi-modal AI models that are generalizable, stable, and interpretable.
\item Verification should be included in the workflow.
\end{itemize}

\section{AI for Test and Verification}

\subsection{Problem Description and Challenges}
\label{subsec:test_verify_problem}

\noindent Test and verification encompass a very broad topic covering a wide range of robustness challenges: (a) design bugs introduced during chip design; (b) manufacturing defects and hardware reliability failures; (c) security vulnerabilities of chips; (d) fragility and security vulnerabilities of AI-based software systems.

Hardware design verification of chips is a major, perhaps the most significant, design challenge today. Pre-silicon verification is used to detect logic design bugs before chips are manufactured. Several industrial reports highlight significant challenges of existing pre-silicon verification practices (e.g., 80\% of design effort may be dedicated to verification, first silicon success lowest since 2004~\cite{verification_trend_report}). The rise of hardware accelerators and multi-chiplet systems makes verification even more challenging. Many design variants and their combinations must now be verified thoroughly and quickly (much quicker and with much more nimble design teams, unlike processors). Major verification challenges include: (a) tremendous efforts in crafting design-specific assertions, tests, properties, and full functional specifications, and (b) critical bug escapes despite “100\% coverage” highlighting the inadequacy of existing approaches~\cite{G-QED}. These challenges get further magnified with AI-generated hardware designs – since AI can generate designs quickly, the relative verification effort increases commensurately.

Bugs that escape pre-silicon verification are detected during post-silicon validation (chips are tested in actual system environments to detect and fix bugs). In addition to logic design bugs, post-silicon validation also targets electrical bugs that are caused by subtle interactions between a design and its electrical state (such as cross-talk, power-supply noise, thermal effects). Post-silicon validation is extremely difficult because the internal signals of a chip are no longer available for controllability and observability, unlike pre-silicon verification.

Beyond design bugs, manufacturing defects are also a major concern. Software engineers typically assume that commodity compute chip hardware — e.g., CPU, GPU, ML accelerator — always works correctly during its lifetime. Large-scale distributed systems in data centers often assume that hardware errors, if any, immediately cause software execution to stop via crashes and hangs, for example, the fail-stop failure assumption. Unfortunately, none of these assumptions holds today. Too many defective compute chips – i.e., with manufacturing defects – are escaping today’s manufacturing tests – at least an order of magnitude more than industrial targets across all compute chip types in data centers. Silent data corruptions (SDCs) caused by test escapes, when left unaddressed, pose a major threat to reliable computing~\cite{silent_data_corruption}.

Reliability failures, largely benign in the past – radiation-induced soft errors, early-life failures, circuit aging – cause unexpected data corruption and expensive system downtimes with severe consequences~\cite{CLEAR}. Business-as-usual manufacturing-time testing, large speed/voltage margins, or expensive redundancy for fault-tolerance aren’t scalable or cost-effective anymore.

The advent of AI-based optimization has dramatically reshaped the landscape of hardware security, presenting unprecedented challenges. The attack surface has expanded to an alarming degree, encompassing not only traditional vulnerabilities within the hardware itself but also novel avenues introduced by the AI integration. Each stage of the design and optimization process, now heavily influenced by AI algorithms, necessitates a rigorously defined formal attack model to effectively identify and mitigate potential threats. This is particularly crucial as the very intelligence intended to optimize design and performance simultaneously introduces new vectors for sophisticated attacks, demanding a proactive and comprehensive re-evaluation of security paradigms. Similarly, the robustness of software systems, which use AI at their core, is becoming highly challenging as well. This is featured by the deep system stack from chips to code, as well as the heterogeneous compute involving non-determinism and complex interactions across hardware platforms, runtime environments, and software ecosystems.

\subsection{Summary of Existing Efforts and Showstoppers}

\noindent LLMs have been explored for hardware design using RTL (such as Verilog) from natural language prompts. Examples of such efforts include~\cite{Chip-Chat,ChipGPT,GPT4AIGChip,Rome,RTLLM,VerilogEval,Verilogcoder,Opl4gpt}. However, these approaches have not yet delivered the expected productivity boost because they suffer from the fundamental design verification bottleneck discussed in Sec.~\ref{subsec:test_verify_problem} – which gets even more severe with hallucinations inherent to today’s AI models and ambiguity inherent to natural language prompts. As a result, frontier models produce RTL designs that are riddled with a prohibitively large number of bugs for complex designs, which is one of the primary reasons that those efforts have not yet delivered the expected productivity boost. 

While HLS techniques, discussed in Sec.~\ref{sec:HLS_LLS}, are promising, HLS also suffers from the design verification bottleneck challenges especially when domain-specific ASICs are targeted. In addition, bugs introduced in HLS pragmas, introduced by human designers or by AI, can be extremely tricky to detect.

AI techniques for design verification and bug fixing (e.g., LLM- or RL-based, such as LLM4SVA discussed in Sec.~\ref{sec:HLS_LLS} and others) are in their infancy today – especially for “difficult” corner-case bugs that get detected by specific input sequences. Existing verification techniques are already highly challenging, and critical bugs escape existing verification flows that are highly time-consuming and resource-intensive in terms of human expertise. Thus, targeting AI techniques that match the quality of existing verification flows isn’t sufficient to ensure design quality and overall design productivity moving forward. Even when bugs are successfully detected, fixing those bugs must also be automated for AI-driven design flows for overall design productivity. Very little progress has been made in the above areas. Needless to say, high-quality training data is crucial for success (e.g., see the discussion on the Chrysalis dataset and corresponding bug injection techniques in Sec.~\ref{sec:HLS_LLS}).

A combination of AI-driven design techniques and recent advances in a new formal verification approach (G-QED~\cite{G-QED}) provides a promising path as presented in~\cite{mitra_presentation} and explained in Sec.~\ref{subsec:test_verify_opportunity}. This approach exploits the unique individual strengths of AI and formal verification in synergistic ways for large productivity and design quality benefits while bypassing the weaknesses of AI approaches for design verification and bug fixing.

Security for hardware and embedded software in the age of AI presents a distinct set of challenges and advantages. Due to the typically lower level of abstraction and the imperative for real-time performance, embedded software often benefits from having no hidden states, making its behavior more predictable and auditable. This inherent transparency, coupled with highly constrained and well-defined use cases, contributes to a more robust security posture compared to general-purpose software. Furthermore, the significantly smaller codebases characteristic of embedded systems reduce the overall attack surface. However, the stringent real-time requirements introduce unique verification hurdles, as security measures must not compromise the system's deterministic and timely operation.

For AI-based optimization and hardware systems, ensuring robust security demands adherence to several critical criteria. Foremost is resilience against a spectrum of malicious inputs, including various poisoning and adversarial attacks that can subtly manipulate system behavior or compromise model integrity. Equally vital is the fidelity of AI model systems to the original design intent, preventing unintended deviations or exploitable backdoors introduced during training or deployment. Finally, meticulous attention must be paid to numerical precision and stability to prevent errors during training that could lead to convergence loss or outright divergence of the training process, ultimately undermining the reliability and security of the entire AI-driven hardware system.

\subsection{Opportunities for AI/EDA Collaboration}
\label{subsec:test_verify_opportunity}

\noindent \textbf{Example 1:}

One of the biggest AI/ML opportunities for chip design is in “creating” highly efficient and robust digital designs with drastic design productivity improvements – e.g., a few days vs. several months. Since verification is the biggest bottleneck in design today, design and verification cannot be treated separately for overall productivity benefits. This creates a unique opportunity to combine AI/ML techniques synergistically with recent breakthroughs in design verification (such as G-QED~\cite{G-QED} and its extensions), which includes non-AI/ML techniques, for an overall AI-driven approach (e.g., AI-Boosted Chip Design or ABCD approach in~\cite{mitra_presentation}). The key idea is to effectively combine these techniques by exploiting their strengths while avoiding their weaknesses. 

To achieve this objective, several deeply technical questions must be answered. A few examples include: 1. How can AI/ML techniques be used to create designs with drastically fewer bugs to start with (before any design verification techniques are applied)? Today’s approaches based on naive natural language prompts create designs riddled with bugs despite small design sizes and no advanced architecture features. 2. How can we verify complex AI/ML-generated designs with minimal human intervention using ideas in~\cite{G-QED} and others? What verification guarantees can be provided, especially for classes of bugs introduced by AI/ML? 3. How can we feed verification results (e.g., counterexamples or bug traces) back to AI/ML for bug fixes? Can we rely on AI/ML-assisted debugging? 4. How do we ensure that bug fixes are correct and don’t significantly degrade the performance and energy efficiency of resulting designs? Can AI/ML and non-AI/ML techniques collaborate on this aspect? 5. How do we ensure that AI/ML understands various architecture features “correctly” and uses them in “appropriate” ways when creating designs?

\noindent \textbf{Example 2:}

Little research literature exists on the effective use of AI for post-silicon validation and debug. Data-driven techniques have been explored for manufacturing testing, mainly from the perspective of reducing cost, e.g., manufacturing test times. However, as discussed in Sec.~\ref{subsec:test_verify_problem}, today’s manufacturing testing practices face major challenges resulting in significant amounts of defective chips escaping into the field. There is an opportunity to explore the role of AI/ML in this context~\cite{silent_data_corruption}. For example, one promising application is identifying potentially defective machines (i.e., machines containing defective chips) in data centers by analyzing system-level signals – uncovering "needle-in-a-haystack" patterns beyond existing rule-based methods. Another direction is using data-driven approaches for in-field testing and error detection – deciding when to schedule them, which hardware blocks to target, or what test content to apply – based on various hardware and software events and on-chip sensor data. Accurate diagnosis from system-level incorrect behaviors and insights obtained from well-designed test experiments can potentially generate high-quality training data for such AI/ML. Finally, understanding how test escape-induced errors impact AI/ML workloads at the application level is critical for ensuring end-to-end robustness.

\noindent \textbf{Example 3:}

Another avenue for deep collaboration between the AI and EDA communities is the use of ML in solvers, provers, analyzers, and verifiers that are integral to EDA tools. Over the last decade, we have seen increasing integration of ML into EDA tools, resulting in dramatic improvement in performance and adaptability of these tools. This is best understood when we notice that solvers (e.g., SAT and SMT solvers), provers (e.g., first-order provers), analyzers (e.g., static analysis tools), and verifiers (e.g., bounded model checkers) are rule-based. Further, traditionally, these systems used ad-hoc heuristics at decision and choice points to optimally sequence, select, and initialize these rules. However, these ad-hoc rules can be replaced with ML-based heuristics, resulting in a novel class of neurosymbolic AI techniques~\cite{ML_and_logic}, that are data-driven and solve the optimization problems associated with rule selection, sequencing, and initialization. Perhaps the best representatives of this approach are Maple* series of SAT solvers that have had a dramatic improvement in the SAT and SMT solver community~\cite{Exponential_recency}, and the benefits of which have translated into far more performant EDA tools today.

Once again, the key insight of this line of work is that there is great benefit in combining ML and verification tools in a symbiotic corrective feedback loop (e.g., reinforcement learning with symbolic feedback~\cite{RLSF}) where the verifier provides corrective feedback to the generative ML and the ML generates formal objects (and data) for the verifier/prover to check~\cite{ML_and_logic}.

\subsection{Future Directions and Recommendations to NSF}
\begin{itemize}
\item A neurosymbolic symbiotic loop between solvers/provers/verifiers and AI: NSF should promote research in neurosymbolic AI techniques that enable the development of ML techniques to improve solvers/provers/verifiers, and in the reverse direction, use solvers/provers/verifiers and domain knowledge to improve the trustworthiness of ML.

\item A neurosymbolic symbiotic loop between synthesizers and AI: AI tools, such as LLMs, are surprisingly good at generating code, proofs, and other formal objects. Having said that, with corrective feedback from provers and verifiers during training and fine-tuning (e.g., RLSF), the quality of their output can be dramatically improved. Hence, NSF should focus on combining synthesizers in corrective feedback loops with AI systems.

\item Neurosymbolic AI tools for HLS and RTL synthesis/verification and bug-fixing: NSF should also promote the use of provers/verifiers as providing symbolic feedback to AI-driven synthesis during inference (in-context symbolic feedback).

\item Neurosymbolic AI tools post-silicon, debug, reliability: NSF should promote similar neurosymbolic AI tools for post-silicon, debug, and reliability.

\item AI software/embedded systems and security – AI systems introduce new attack surfaces that need to be addressed.
\end{itemize}

\section{Acknowledgments}
\noindent This workshop was sponsored by the NSF \#2436036, Google, AMD, and The Institute for Learning-enabled Optimization at Scale (TILOS). We thank the speakers and panelists for their insightful contributions, including Deming Chen, Jeff Dean, Vijay Ganesh, Aditya Grover, Farinaz Koushanfar, Koen Lampaert, Yingyan (Celine) Lin, Yong Liu, Igor Markov, Subhasish Mitra, Bryan Perozzi, Pranay Prakash, Ruchir Puri, Mark Ren, Shobha Vasudevan, Yusu Wang, and Lingming Zhang. We also thank the workshop organizers for their leadership and coordination, including Jason Cong, Sergio Guadarrama, Stefanie Jegelka, David Z. Pan, and Yizhou Sun. We gratefully acknowledge the support of the NSF program directors, including Raj Acharya, Sankar Basu, and Sharon Hu. Finally, we thank the student volunteers who assisted with the organization, including Zijian Ding, Weikai Li, Zongyue Qin, and Derek Xu.

\bibliographystyle{IEEEtran}
\bibliography{reference}


\begin{IEEEbiography}[{\includegraphics[width=1in,height=1.25in,clip,keepaspectratio]{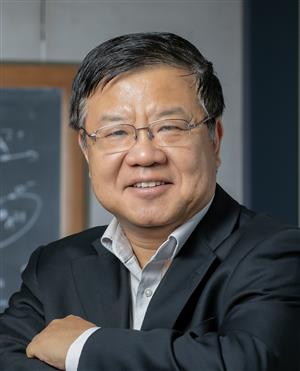}}]{Deming Chen}
(Fellow, IEEE) received the Ph.D. degree in computer science
from UCLA in 2005. He is a Abel Bliss Professor at the Grainger College of Engineering, University of Illinois Urbana-Champaign. He has published more than 300 research papers, received ten Best Paper Awards, two ACM/SIGDA TCFPGA Hall-of-Fame Paper Awards, five Best Poster Awards, and delivered more than 170 invited talks, including over 20 keynote and distinguished lectures. His research interests include machine learning and AI, system-level design methodologies, hybrid cloud systems, security and confidential computing, and reconfigurable and heterogeneous computing. He is an ACM Fellow and previously served as the Editor-in-Chief for ACM Transactions on Reconfigurable Technology and Systems (TRETS). Under his leadership, the impact factor of ACM TRETS has increased by 3.8 times. He serves as the Illinois Director for the IBM-Illinois Discovery Accelerator Institute and the Director of the AMD Center of Excellence. Additionally, he has been involved in several startup companies, including AutoESL and Inspirit IoT.
\end{IEEEbiography}

\begin{IEEEbiography}[{\includegraphics[width=1in,height=1.25in,clip,keepaspectratio]{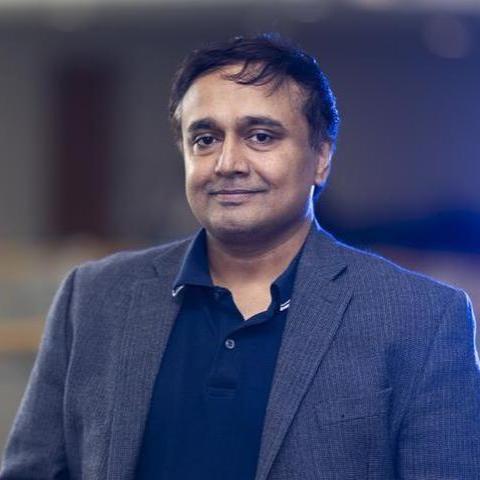}}]{Vijay Ganesh}
received the Ph.D. degree in computer science from Stanford University in 2007. He is a Professor of computer science and an Associate Director of the Institute for Data Engineering and Science (IDEaS), Georgia Institute of Technology. Additionally, he is the Co-Founder and the Director of the AI for Math Center, Georgia Tech, and the Co-Founder of the Centre for Mathematical AI, Fields Institute. Prior to joining Georgia Tech, in 2023, He was a Professor at the University of Waterloo from 2012 to 2023, the Co-Director of the Waterloo AI Institute from 2021 to 2023, and a Research Scientist at MIT from 2007 to 2012. His primary research area is the theory and practice of SAT/SMT solvers, combinations of machine learning and automated reasoning, and their application in neurosymbolic AI, software engineering, security, mathematics, and physics. He has led the development of many SAT/SMT solvers, most notably, STP, Z3str family of string solvers, Z3-alpha, MapleSAT, AlphaMapleSAT, and MathCheck. He also leads the development of several neurosymbolic AI tools and aimed at mathematics, physics, and software engineering. On the theoretical side, he works on topics in mathematical logic and proof complexity. For his research, he has won over 35 awards, honors, and medals, including the ACM Impact Paper Award at ISSTA 2019, the ACM Test of Time Award at CCS 2016, and the Ten-Year Most Influential Paper citation at DATE 2008.
\end{IEEEbiography}

\vspace{-1mm}

\begin{IEEEbiography}[{\includegraphics[width=1in,height=1.25in,clip,keepaspectratio]{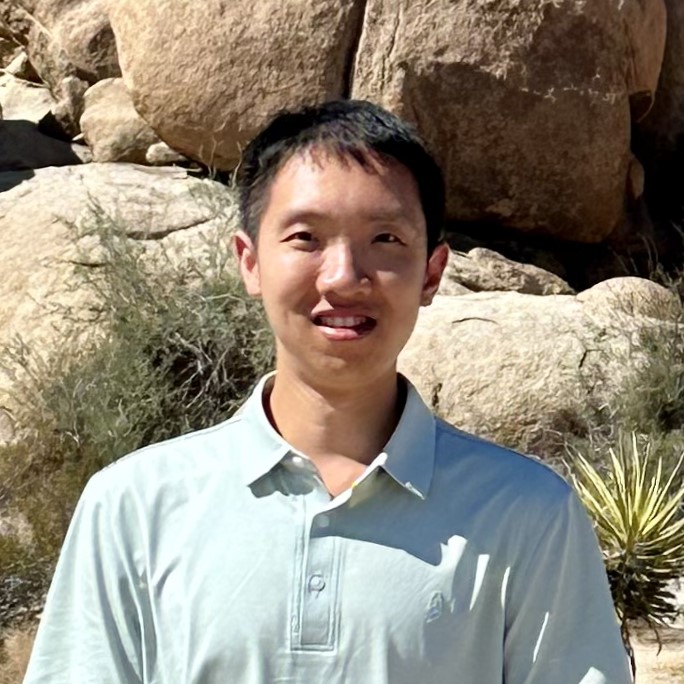}}]{Weikai Li}
received the bachelor’s degree in computer science and technology from Tsinghua University in 2023. He is currently pursuing the Ph.D. degree with the Computer Science Department, UCLA, under the supervision of Prof. Yizhou Sun. His research focuses on knowledge-enhanced AI for chip design, improving the capability and generalizability of graph neural networks (GNNs) and large language models (LLMs) for chip design.
\end{IEEEbiography}

\begin{IEEEbiography}[{\includegraphics[width=1in,height=1.25in,clip,keepaspectratio]{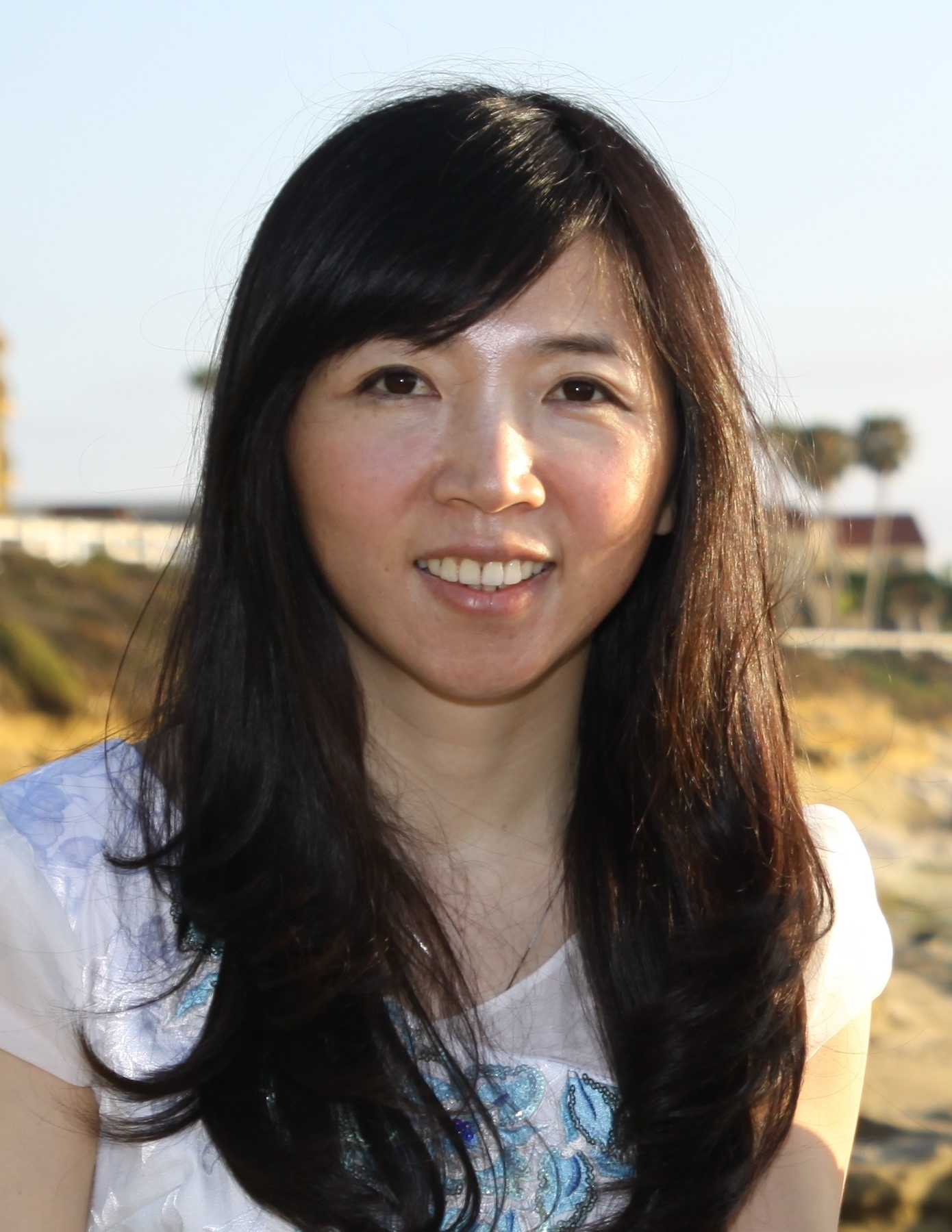}}]{Yingyan (Celine) Lin}
is an Associate Professor at the School of Computer
Science and the Co-Director of the Center for Advancing Responsible Computing, Georgia Institute of Technology. She leads the Efficient and Intelligent Computing (EIC) Laboratory, focusing on efficient AI solutions, with an emphasis on cross-layer algorithm-hardware co-design and agentic AI techniques for hardware design. Her group’s research has earned various recognitions, including First Place at the ACM SIGDA University Demonstration at DAC 2022, First Place at the ACM/IEEE TinyML Design Contest at ICCAD 2022, and selection as the IEEE Micro Top Pick of 2023 for “the most significant research papers in computer architecture based on novelty and potential for long-term impact.” Additionally, their work has been spotlighted at ICLR in 2020, 2021, and 2025, and NeurIPS 2025, presented as an oral paper at ECCV 2024, and received the Best Paper Award at MICRO 2024. She is the PC Co-Chair for both MLSys 2025 and the IEEE MICRO Top Picks 2026.
\end{IEEEbiography}

\begin{IEEEbiography}[{\includegraphics[width=1in,height=1.25in,clip,keepaspectratio]{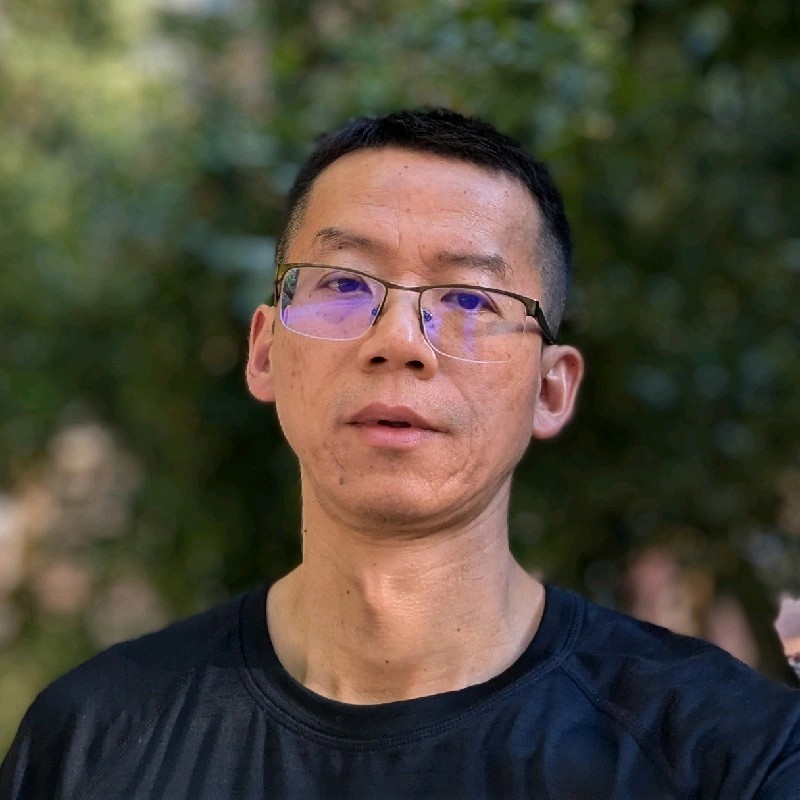}}]{Yong Liu}
received the M.S. degree in computer science from Fudan University. He is the Senior Group Director at Cadence Design Systems, where he leads research and development teams on large-scale data analytics, artificial intelligence, LLM infrastructure, and applications, including JedAI and ChipGPT. He is an Experienced Architect in electronic design automation (EDA), AI hardware–software co-design, and AI-driven data analytics. Prior to joining Cadence, he contributed to AI chip development and formal verification tooling across multiple technology companies. He is the Technical Program Committee (TPC) Co-Chair of IEEE International Conference on LLM-Aided Design.
\end{IEEEbiography}

\begin{IEEEbiography}[{\includegraphics[width=1in,height=1.25in,clip,keepaspectratio]{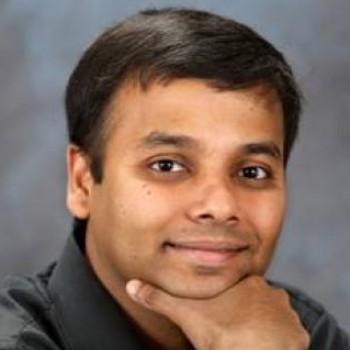}}]{Subhasish Mitra}
(Fellow, IEEE) holds the William E. Ayer Endowed Chair Professorship at the EECS Department, Stanford University. He directs the Stanford Robust Systems Group, serves on the leadership team of the Microelectronics Commons AI Hardware Hub funded by the US CHIPS and Science Act, leads the Computation Focus Area of the Stanford SystemX Alliance, and is the Associate Chair (Faculty Affairs) of Computer Science. His research ranges across robust computing, nanosystems, electronic design automation (EDA), and neurosciences. He is a fellow of ACM. His honors include the Harry H. Goode Memorial Award (by IEEE Computer Society), Newton Technical Impact Award in EDA (test-of-time honor by ACM SIGDA and IEEE CEDA), the University Researcher Award (by Semiconductor Industry Association and Semiconductor Research Corporation, the EDAA Achievement Award (by European Design and Automation Association), the Intel Achievement Award (Intel’s highest honor), and the Distinguished Alumnus Award from the Indian Institute of Technology, Kharagpur.
\end{IEEEbiography}

\begin{IEEEbiography}[{\includegraphics[width=1in,height=1.25in,clip,keepaspectratio]{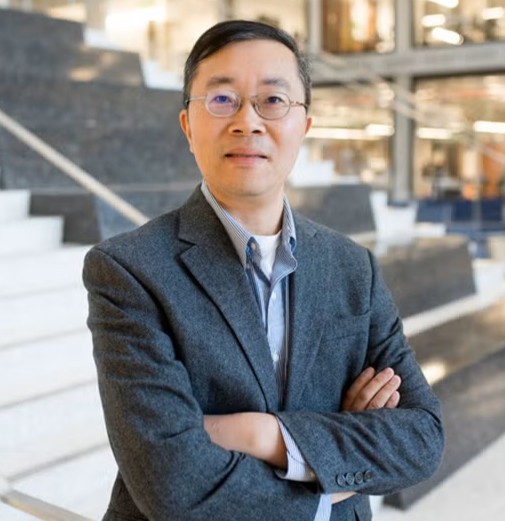}}]{David Z. Pan}
(Fellow, IEEE) received the B.S. degree from Peking University and the M.S. and Ph.D. degrees from the University of California at Los Angeles. He is a Professor and a holder of the Silicon Laboratories Endowed Chair at the ECE Department, UT Austin. He has published over 500 refereed papers at top EDA/chips/architecture and machine learning venues. His research interests are mainly focused on electronic design automation (EDA), including how to leverage AI/ML for EDA in physical design and design for manufacturing, as well as synergistic AI and IC co-optimizations. He is a fellow of ACM and SPIE. He has served as the Program Chair for DAC in 2024 and ICCAD in 2018, and the two flagship conferences in EDA. He has received many awards, including the SRC Technical Excellence Award, the DAC Top 10 Author in Fifth Decade, the DAC Prolific Author Award, the ASP-DAC Frequently Cited Author Award, the 21 Best Paper Awards, and over 20 additional Best Paper candidates/finalists, and a number of international CAD contest awards, among others.
\end{IEEEbiography}

\begin{IEEEbiography}[{\includegraphics[width=1in,height=1.25in,clip,keepaspectratio]{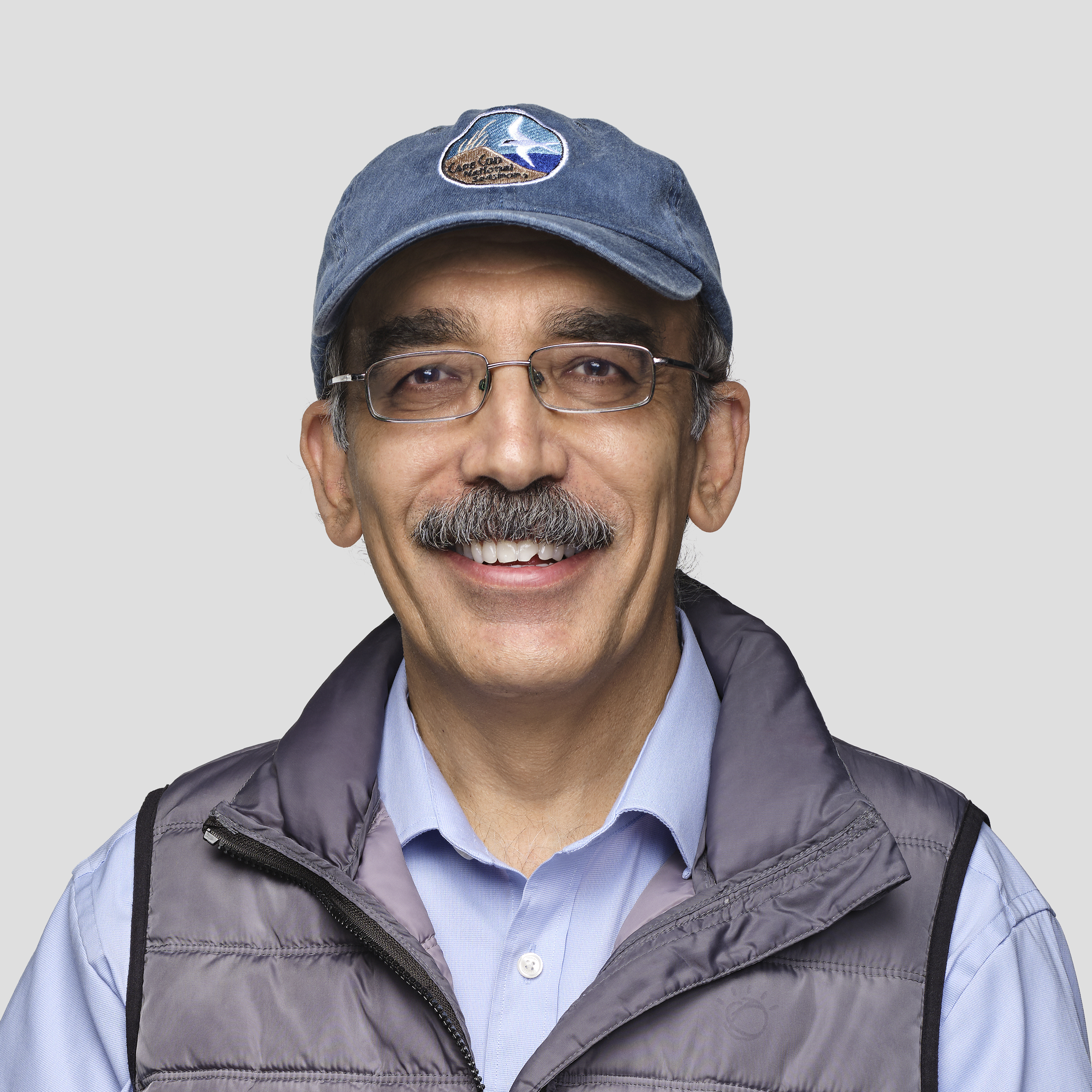}}]{Ruchir Puri}
(Fellow, IEEE) is the Chief Scientist of IBM Research, an IBM Fellow, and the Vice-President of IBM Corporate Technology. He led IBM Watson as the CTO and the Chief Architect from 2016 to 2019 and has held various technical, research, and engineering leadership roles across IBM’s AI and Research businesses. He has been an ACM Distinguished Speaker, an IEEE Distinguished Lecturer, and was awarded the 2014 Asian American Engineer of the Year. He has been an Adjunct Professor at Columbia University, NY, and a Visiting Scientist at Stanford University, CA. He was honored with the John Von-Neumann Chair at the Institute of Discrete Mathematics at Bonn University, Germany. He is an Inventor of over 75 U.S. patents and has authored over 150 scientific publications. He is the Founder of the IEEE LLM-Aided Design Conference. He was a recipient of the Distinguished Alumnus Award from Indian Institute of Technology (IIT), Kanpur.
\end{IEEEbiography}

\begin{IEEEbiography}[{\includegraphics[width=1in,height=1.25in,clip,keepaspectratio]{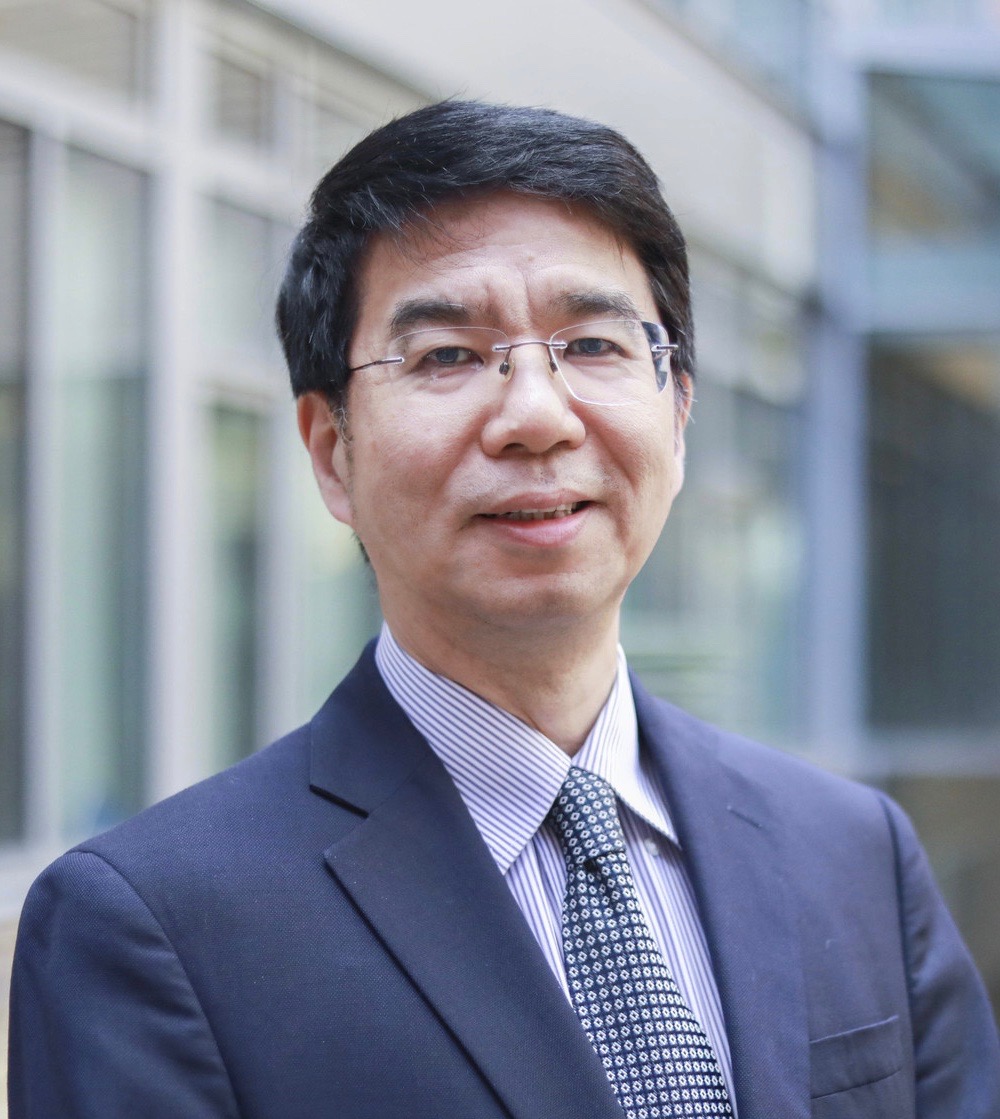}}]{Jason Cong}
(Fellow, IEEE) is the Volgenau Chair for Engineering Excellence Professor at the Computer Science Department, UCLA, with joint appointment from the Electrical and Computer Engineering Department. He is the Director of the Center for Domain-Specific Computing (CDSC) and the Director of the VLSI Architecture, Synthesis, and Technology (VAST) Laboratory. His research interests include novel architectures and compilation for customizable computing, synthesis of VLSI circuits and systems, and quantum computing. He is a member of the National Academy of Engineering, the American Academy of Arts and Sciences, and a fellow of ACM and the National Academy of Inventors. He was a recipient of the SIA University Research Award and the EDAA Achievement Award. He was honored for the IEEE Robert N. Noyce Medal “for fundamental contributions to electronic design automation and FPGA design methods” in 2022, the Phil Kaufman Award, the highest recognition in EDA, in 2024, and the ACM Chuck Thacker Breakthrough Award in 2025 “for fundamental contributions to the design and automation of field-programmable systems and customizable computing.”
\end{IEEEbiography}

\begin{IEEEbiography}[{\includegraphics[width=1in,height=1.25in,clip,keepaspectratio]{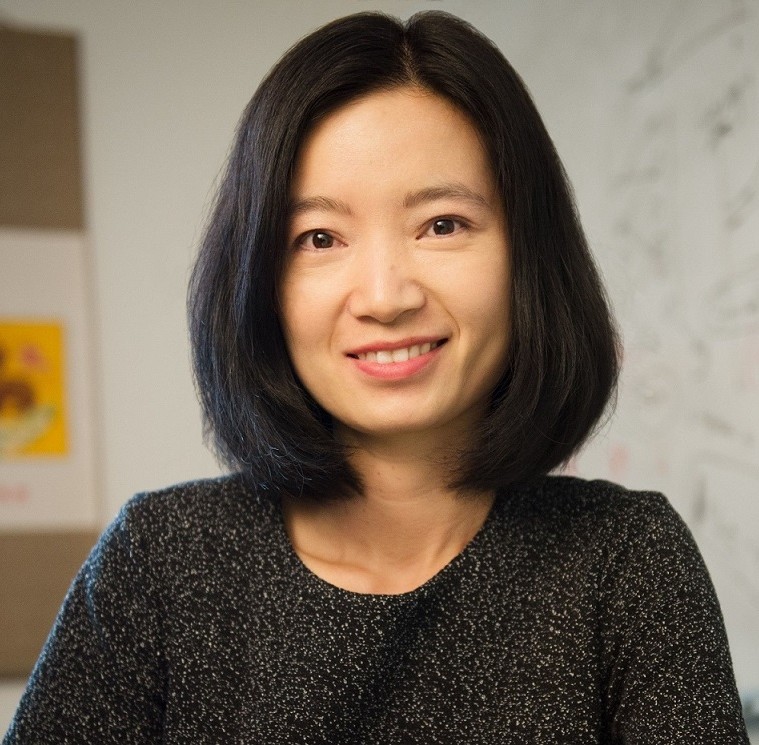}}]{Yizhou Sun}
(Senior Member, IEEE) received the Ph.D. degree in computer science from the University of Illinois at Urbana-Champaign in 2012. She is a Professor at the Department of Computer Science, UCLA, and an Amazon Scholar. She is a Pioneer Researcher in mining heterogeneous information networks, with a recent focus on deep learning on graphs, AI for chip design, AI for science, and neuro-symbolic reasoning. She has over 200 publications in books, journals, and major conferences. Tutorials on her research have been given in many premier conferences. She was a ACM Distinguished Member. She was a recipient of multiple Best Paper Awards, the ACM SIGKDD Doctoral Dissertation Award, the Yahoo Academic Career Enhancement (ACE) Award, the NSF CAREER Award, the CS@ILLINOIS Distinguished Educator Award, the Amazon Research Awards (twice), the Okawa Foundation Research Award, the VLDB Test of Time Award, the WSDM Test of Time Award, the IEEE AI’s 10 to Watch, and the SDM/IBM Faculty Award. She is the General Co-Chair of SIGKDD 2023, the PC Co-Chair of ICLR 2024, and the PC Co-Chair of SIGKDD 2025.
\end{IEEEbiography}

\vfill

\end{document}